\documentclass[runningheads]{llncs}
\usepackage{graphicx}
\usepackage{amsmath,amssymb} 
\usepackage{color}
\usepackage[width=122mm,left=12mm,paperwidth=146mm,height=193mm,top=12mm,paperheight=217mm]{geometry}

\usepackage{algorithmic}
\usepackage{makecell}
\usepackage{colortbl}
\definecolor{Gray}{gray}{0.9}
\usepackage{multirow}
\usepackage{booktabs}
\usepackage[ruled,linesnumbered]{algorithm2e}

\makeatletter 
\def\thickhline{%
  \noalign{\ifnum0=`}\fi\hrule \@height \thickarrayrulewidth \futurelet
   \reserved@a\@xthickhline}
\def\@xthickhline{\ifx\reserved@a\thickhline
               \vskip\doublerulesep
               \vskip-\thickarrayrulewidth
             \fi
      \ifnum0=`{\fi}}
\makeatother

\newlength{\thickarrayrulewidth}
\usepackage{hyperref}
\hypersetup{
            colorlinks=true,
            linkcolor=blue,
            anchorcolor=blue,
            citecolor=blue}

\begin{document}
\pagestyle{headings}
\mainmatter
\def\ECCV16SubNumber{***}  

\title{Progressive Hard-case Mining across Pyramid Levels for Object Detection} 

\author{Binghong Wu \and
        Yehui Yang* \and
        Dalu Yang \and
        Junde Wu  \and
        XiaoRong Wang \and
        Haifeng Huang  \and
        Lei Wang  \and
        Yanwu Xu}
\institute{Intelligent Healthcare Unit,  Baidu Inc. \\
*Corresponding author and project leader: yangyehuisw@126.com}

\maketitle

\begin{abstract}
    In object detection, multi-level prediction (e.g., FPN) and reweighting skills (e.g., focal loss) have drastically improved one-stage detector performance. However, the synergy between these two techniques is not fully explored in a unified framework. We find that, during training, the one-stage detector's optimization is not only restricted to the static hard-case mining loss (\emph{gradient drift}), but also suffered from the diverse positive samples' proportions split by different pyramid levels (\emph{level discrepancy}). Under this concern, we propose Hierarchical Progressive Focus (HPF) consisting of two key designs: 1) \emph{progressive focus}, a more flexible hard-case mining setting calculated adaptive to the convergence progress, 2) \emph{hierarchical sampling}, automatically generating a set of progressive focus for level-specific target optimization. Based on focal loss with ATSS-R50, our approach achieves 40.5 AP, surpassing the state-of-the-art QFL (Quality Focal Loss, 39.9 AP) and VFL (Varifocal Loss, 40.1 AP). Our best model achieves \textbf{55.1} AP on COCO \emph{test-dev}, obtaining excellent results with only a typical training setting. Moreover, as a plug-and-play scheme, HPF can cooperate well with recent advances, providing a stable performance improvement on \textbf{9} mainstream detectors.
\end{abstract}

\section{Introduction}\label{sec:intro}
    One-stage object detectors are popular in practical applications because of their higher efficiency and lower deployment cost than multi-stage detectors \cite{zou2019object}. Recently, one-stage detectors have gradually caught up with multi-stage detectors in terms of accuracy, benefitting from advances in model architectures \cite{liu2016ssd,lin2017focal}, loss functions \cite{zhang2021varifocalnet,rezatofighi2019generalized}, target assignment strategies \cite{tian2019fcos,zhang2020bridging}, etc. 

    \begin{figure}[t]
        \centering
        \includegraphics[width=1\columnwidth]{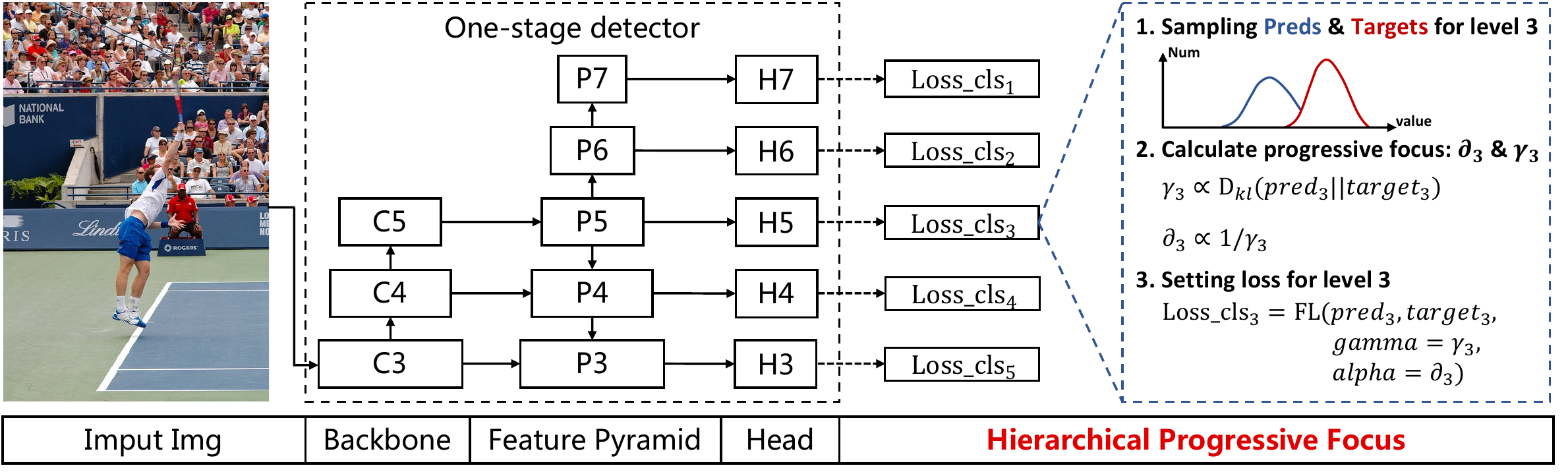}
        \caption{HPF optimizes each level with a progressive mining strategy, based on \textbf{different} ${\rm D}_{kl}$ (KL distance) at which level they are (details is shown as Eq. \ref{form 4} and Eq. \ref{form 5}).}
        \label{fig_1}
    \end{figure}

    Of these recent advances, two techniques provide vital improvements for one-stage detectors: 1) Feature Pyramid Network (FPN) \cite{lin2017feature}, which provides dense candidates with diverse receptive fields, and improves the performance by solving the mismatch between the receptive field and the object scale. 2) Focal loss \cite{lin2017focal}, which reweights all potential proposals according to the margin between the target and its probability, improves the performance by enabling online hard-case mining. These two techniques have been widely used in leading studies, including anchor-based and anchor-free detectors \cite{lin2017focal,tian2019fcos,zhu2019feature,zhang2020bridging,tan2020efficientdet,kong2020foveabox}.

    Despite their respective contributions, there still exist two limitations when they are both explored in a unified framework. 1) \emph{level discrepancy}: FPN and focal loss are not mutually improved. As a divide-and-conquer solution, FPN captures objects of different scales into different pyramid levels \cite{chen2021you,zhang2020bridging}. This behavior results in \emph{level discrepancy}, where the proportions of positive samples (and hard cases) vary among pyramid levels, requiring level-specific optimization. However, focal loss only has two fixed parameters as it is designed from a \emph{global} optimization perspective, which may be suitable for partial pyramid levels, but not all levels. We provide a detailed evaluation in Fig. \ref{fig2} and a statistical analysis in $\S$\ref{sec:levelImbalance}. 2) \emph{gradient drift}: the static settings of focal loss cannot satisfy the training procedure all the time. Focal loss sets two \emph{fixed} parameters to enhance the hard case's position in optimization. However, easy cases may gradually dominate the optimization with training going on, when there are still many hard samples that have not been correctly classified. We refer it as \emph{gradient drift} in Fig. \ref{fig_3} (a) and perform a detailed analysis in $\S$\ref{sec:staticMining}.
    
    Compared with one-stage detectors, multi-stage detectors can mitigate optimization difficulties by applying a hierarchical mechanism \cite{cai2018cascade}.
    At inference, such a framework can progressively refine results step by step \cite{ren2015faster,cai2018cascade,pang2019libra,chen2019hybrid}.
    During training, it alleviates the imbalance by the \emph{differentiated} resampling mechanisms for different stages (i.e., Cascade R-CNN\cite{cai2018cascade}, HTC \cite{chen2019hybrid}). 
    A natural question arises: can we utilize a similar hierarchical optimization scheme on one-stage detectors to promote the performance, through the lens of divide-and-conquer optimization?
    
    Inspired by multi-stage detectors, we propose Hierarchical Progressive Focus (HPF), which can improve the synergy between focal loss and FPN by two designs: 1) \emph{progressive focus}, which can adaptively obtain substantial gradients from hard cases, avoiding the gradient drift problem; 2) \emph{hierarchical sampling}, which can automatically generate a set of progressive focus for various pyramid levels, mediating the influence of level discrepancy.

    Our main contributions are:
    \begin{itemize}
        \item To the best of our knowledge, we are the first to experimentally show that the level discrepancy limits the performance of one-stage detectors to some extent. Therefore, we design HPF, promoting ATSS \cite{zhang2020bridging} by $\rm \sim1.5 AP$ without any network structure refinement. 

	    \item As a plug-and-play freebie, HPF can guarantee a stable improvement on \textbf{9} representative detectors. Results show that HPF can cooperate well with various target assignment strategies, anchor definitions, and other extensions of focal loss.

        \item Without bells and whistles, our best model achieves \textbf{55.1} AP on MS COCO \emph{test-dev} (comparable performance with EfficientDet-D7x, but requiring only 5$\%$ training resource and  4$\%$ training epochs) is easy to follow and by far the SOTA in one-stage detectors.
    \end{itemize}

\section{Related Work}\label{sec:relatedWork}
    \subsection{Optimization in Multi-stage Detectors}\label{sec:relatedMultiStage}
    Benefitting from its inherent structure, multi-stage detectors can mitigate optimization difficulties stage by stage. Originating from the sliding-window approaches \cite{dalal2005histograms,felzenszwalb2010cascade}, the two-stage detector inherits the paradigm of locating first and refining later, achieving a better performance \cite{girshick2014rich,girshick2015fast,ren2015faster}. From sliding-window methods to selective search, and then to Region Proposal Network, it is evident that the development of ROI (region of interest) extraction promotes performance significantly. From our point of view, the Region Proposal Network \cite{ren2015faster} can be regarded as a data-driven resampling scheduler according to the matching quality, discarding a large amount of low-quality negative samples firstly to alleviate the target imbalance during optimization. Additionally, the head of multi-stage detectors \cite{cai2018cascade,chen2019hybrid} can eliminate the low-quality predictions step by step for solving the quality mismatch during both optimization and inference. Recently, there are still many further studies on the optimization method with various novel perspectives. In the training procedure, IoU-balanced sampling \cite{pang2019libra} creates a better target distribution, assigning samples based on matching quality in each pyramid level. Besides, many studies \cite{song2020revisiting,zhu2021cpm} try to decouple the recognition and localization task, improving the performance by separately sampling decoupled features. For high-quality object detection, the IoU-guided NMS (Non-Maximum Suppression) method can bring the prediction of localization quality into NMS for a better post-processing calibration \cite{jiang2018acquisition}.

    \subsection{Optimization in One-stage Detectors}\label{sec:relatedOneStage}
    The studies on loss design make one-stage detectors comparable to two-stage detectors. Inspired by OHEM, focal loss \cite{lin2017focal} is proposed to make the model focus more on hard cases by extremely down-weighting the easy cases' losses and slightly reducing the weights of hard cases, which has been widely used recently. With much more dense candidates provided by multi-level architectures \cite{liu2016ssd,lin2017feature,redmon2017yolo9000}, reweighting has been regarded as a critical approach for a better performance stably. Until now, such a design can also keep vital contributions on various detectors, implicitly providing a solid foundation for the explosive growth of anchor-free detectors \cite{zhu2019feature,tian2019fcos}. Furthermore, many studies take such an idea into account much more deeply. For example, a penalty-reduced pixel-wise loss based on focal loss has been proposed to optimize the prediction of center points in CenterNet \cite{zhou2019objects}, and a gradient harmonizing mechanism \cite{li2019gradient} is proposed to ensure optimization robustness during training.
    
    Recently, there occurred many focal loss extensions for better quality estimation. For example, GFL merges the localization estimation into the classification branch, and extends focal loss compatible with continuous space \cite{li2020generalized,li2021generalized}. Based on IoU-aware class scores, VFL \cite{zhang2021varifocalnet} weights positive and negative samples asymmetrically, leading to an excellent performance as well. Although so many extensions have been derived, previous studies just focused on pursuing a better target definition rather than the mining approach. As a result, the global constant mining settings continue to be used today, leading to a fixed mining criterion through all levels and all training iterations, even after hard cases gradually reduced with the training proceeding.

    \subsection{Optimization by Target Assignment}\label{sec:relatedAssign}
    During optimization, target assignment strategy affects target distribution deeply. From a unified perspective on both anchor-based and anchor-free detectors, the issue of target assignment has been gradually noticed in recent years \cite{ke2020multiple}. ATSS \cite{zhang2020bridging} proved that the critical difference between RetinaNet and FCOS is the target assignment strategy, resulting in a performance gap by applying different target distributions. Correspondingly, they design a novel strategy with dynamic IoU thresholds on pyramid levels for better performance. Besides, AutoAssign \cite{zhu2020autoassign} improves the performance by an entirely data-driven method. Regrading it as an online combinatorial optimization problem, PAA \cite{kim2020probabilistic} and OTA \cite{ge2021ota} design their specific objective function and find a suitable online strategy separately for each iteration. These strategies provide various target distributions for remarkable performance improvement. Does not conflict with these existing works, our method delves into the detector optimization level by level, and improves the performance with a better hard-case mining approach.

\section{Problem Formulation}\label{sec:motivation}
	In this section, we first revisit the original focal loss and its limitations, i.e., the gradient drift problem. Next, we provide qualitative and quantitative analyses on how FPN leads to the level discrepancy and its influence on performance.
	
    \subsection{Gradient Drift}\label{sec:staticMining}
    Same to OHEM \cite{shrivastava2016training}, we define hard cases as the samples that produce relatively large losses. For binary classification, focal loss is defined as Eq. \ref{ori focal}:

    \begin{align}
        \label{ori focal}
        {\rm FL}(p_i, y_i)=\left\{
        \begin{aligned}
            &-\alpha (1-p_i)^\gamma {\rm log}(p_i), &      &  y_i=1 \\
            &-(1-\alpha)p_i^\gamma {\rm log}(1-p_i), &      &  y_i=0.
        \end{aligned}
        \right.
    \end{align}%

    In Eq. \ref{ori focal}, $p_i$ indicates the model's result activated by the sigmoid function. $y_i \in \{0,1\}$ is the assigned label powered by a specific assignment strategy for $p_i$. In the original focal loss paper \cite{lin2017focal}, $\alpha$ is a well-tuned constant value to keep the gradient balance between the positive and the negative samples, and $\gamma$ balances the optimization between easy and hard samples, through a resampling approach according to the margin between the assigned label and the current prediction. 
    
    The specific setting of $\alpha$ and $\gamma$ affects the performance significantly \cite{lin2017focal}. With appropriate $\alpha$ and $\gamma$, the gradients of massive easy and negative samples are compressed well. In addition, as a setup attached to $\gamma$, a lower $\alpha$ always corresponds to a higher $\gamma$, modulating the focus on positive samples with more easy negatives hugely down-weighted, and keeping the balance among all losses.

    \begin{figure}[tbp]
        \centering
        \includegraphics[width=1\columnwidth]{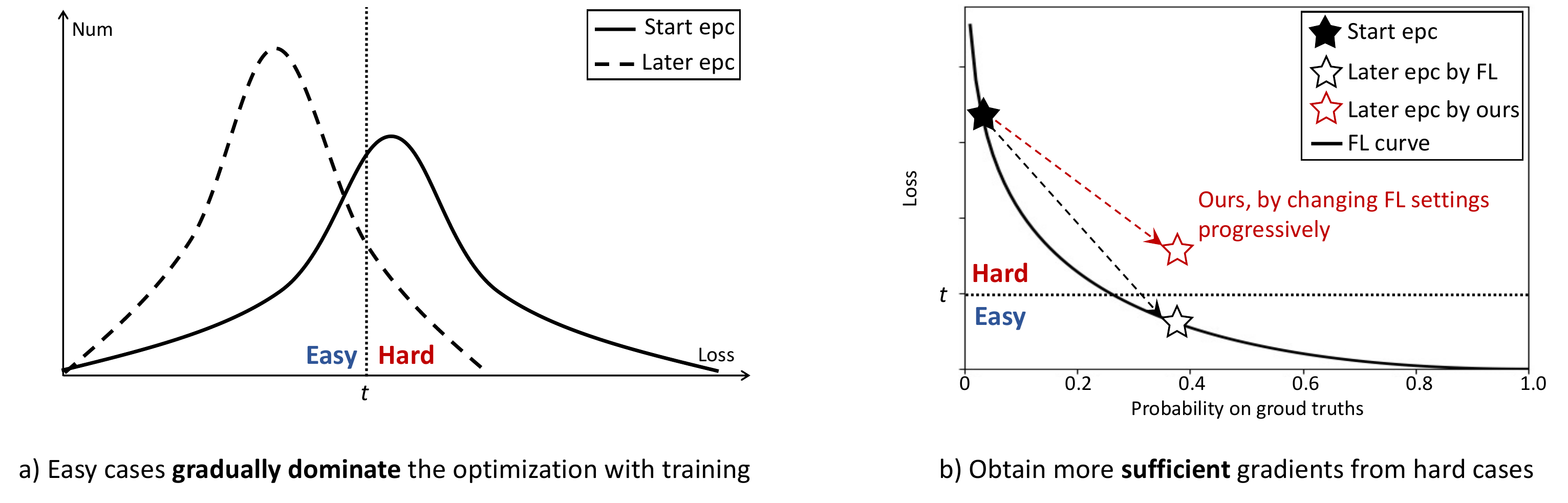}
        \caption{We sketch \emph{easy}/\emph{hard} with a loss threshold \emph{t}. \textbf{a}) Gradient drift phenomenon: Hard cases initially provide most gradients, but easy cases gradually dominate the optimization with training. However, there are still many samples that have not been correctly classified. \textbf{b}) Compared with the conventional, Progressive Focus (described in $\S$\ref{sec:factor}) can progressively fine-tune the loss's mining settings during training, obtaining more focus from hard cases to guide a better optimization.}
        \label{fig_3}
    \end{figure}
    
    \textbf{Limitations.} The static $\alpha$ and $\gamma$ may restrict the optimization, although they have been well-tuned. We draw our concern as Fig. \ref{fig_3} (a): the optimization is guided by hard cases at the early stage but \emph{progressively drifts} when the training goes on, for many cases are harder at the beginning while relatively easier in the latter.  For example, easy cases still dominate the later-stage optimization, shown as the dotted curve.

    Correspondingly, Fig. \ref{fig_3} (b) illustrates an expected optimization with a better loss setting, which can obtain more efficient gradients from hard cases. As a result, we design a progressive hard-case mining approach to finetune the optimization focus along with the training. As shown in $\S$\ref{sec:factor}, our approach can be described in the following Eq. \ref{form 4} and \ref{form 5}: we propose a bootstrap design on $\alpha_{ad}$ and $\gamma_{ad}$, progressively transferring the focus according to the current situation.

    \subsection{Level Discrepancy}\label{sec:levelImbalance}
    As shown in Fig. \ref{fig_1}, multi-level prediction is widely used in one-stage detectors, drastically improving performance. 
	We analyze various detectors without loss of generality to verify the relations between \emph{level discrepancy} and the performance, including RetinaNet \cite{lin2017focal}, ATSS \cite{zhang2020bridging}, FCOS \cite{tian2019fcos}, and VFNet \cite{zhang2021varifocalnet}.

    \textbf{Experiment settings.} We record the proportions of positive samples for each level and evaluate their final performance independently. All experiments are conducted on MS COCO \cite{lin2014microsoft} dataset, containing 115K images (\emph{trainval35k}) for training and 5K images (\emph{minival}) for evaluation. All detectors are equipped with the same backbone ResNet-50. Also, all detectors adopt the same multi-level prediction architecture: shown as Fig. \ref{fig_1}, FPN obtains C3 to C5 from the backbone and generates a series of feature maps named P3 to P7 for a series of hierarchical predictions.

    \begin{figure*}[t]
        \centering
        \includegraphics[width=1.0\textwidth]{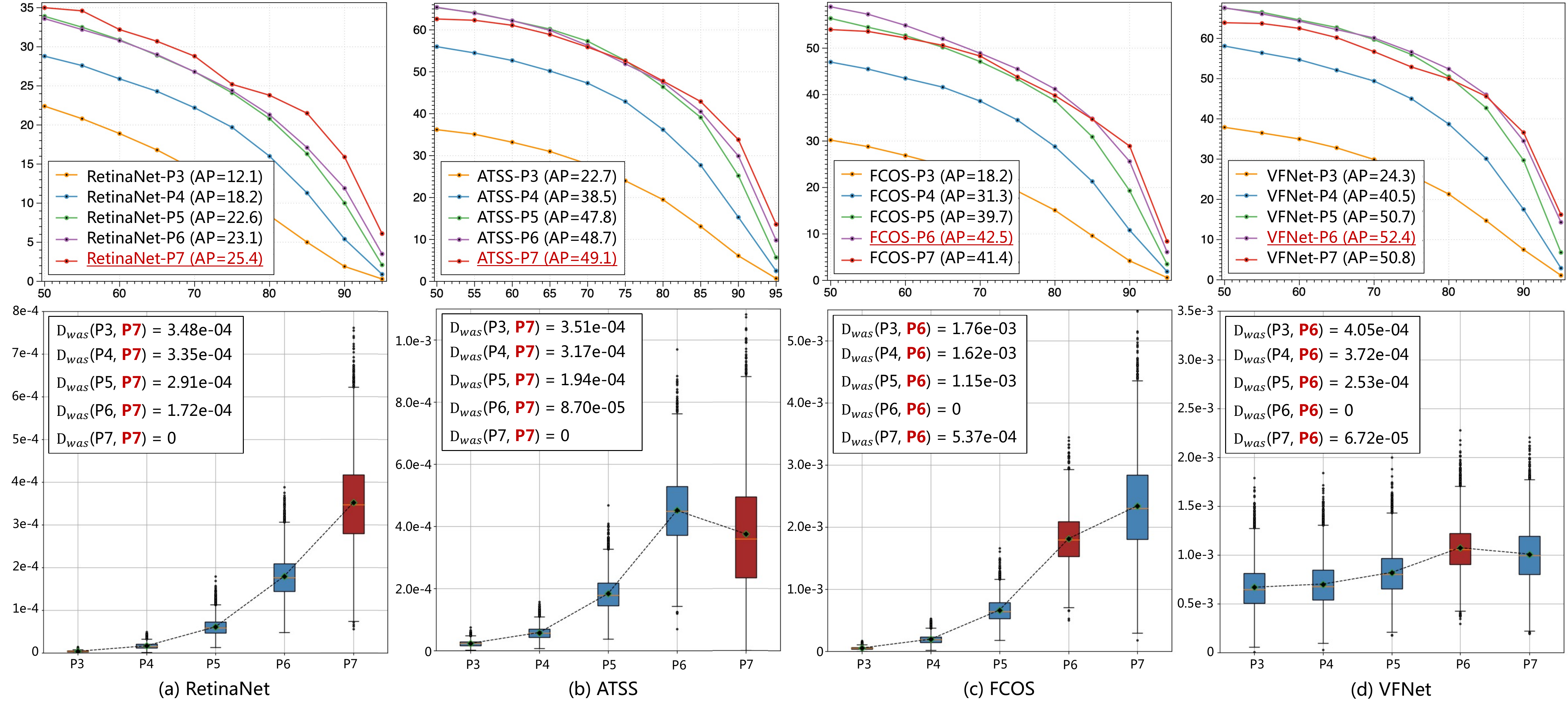}
        \caption{Level discrepancy and performance: P3 to P7 indicate different level predictions. \textbf{1}) We draw level performance as AP-IoU curves at the first line. \textbf{2}) The proportions of positive samples are recorded in each training iteration, and summarized as box plots at the second line. (Detailedly, the height of boxes reflects the variance of the percentages, and the lined dots indicate mean values.) \textbf{3}) Furthermore, we also provide the mAP metrics for all levels and calculate the Wasserstein distances $\rm D_{was}$ from the best level to others.}
        \label{fig2}
    \end{figure*}
    
    \textbf{Evaluation approach.} We conduct statistical analysis within one training epoch: during each iteration, the proportion of positive samples split by pyramid levels is recorded and summarized as the box plots in Fig. \ref{fig2}. Furthermore, we quantitatively evaluate the distribution distance by calculating the Wasserstein distance $\rm D_{was}$ from the best level to others. For level-wise evaluation, we extract a series of subsets from \emph{minival}, which only contains the \emph{level-assigned} targets according to their assignment strategy. Then, we evaluate the level performance with various IoU threshold, and draw AP-IoU curves at the first line in Fig. \ref{fig2}.

    \textbf{Analysis.} We draw the following conclusion based on the results of four detectors in Fig. \ref{fig2}. \emph{Firstly}, the level discrepancy exists in multi-level optimization, shown as box plots in Fig. \ref{fig2}. \emph{Secondly}, each level has different performance correspondingly. Based on the two findings, we believe that focal loss does have an optimization preference for partial levels. As a result, we provide a hierarchical sampling approach, described in $\S$\ref{sec:hieSample}.

\section{Method}\label{sec:method}

    We propose HPF for a better multi-level optimization based on these two concerns. The proposed method consists of two key designs: 1) \emph{progressive focus}, which adjusts the hard-case mining settings progressively, along with the prediction quality (KL distance) of positive samples during training, and 2) \emph{hierarchical sampling}, which can take better care of each layer's target distribution to further improve performance. In addition, we remind that hierarchical sampling \emph{does not independently work} without the progressive focus, for $\alpha$ and $\gamma$ are the same constant for all levels in the original focal loss function.

    \subsection{Progressive Focus}\label{sec:factor}
	Unlike QFL (Quality Focal Loss) \cite{li2020generalized} and VFL (Varifocal Loss) \cite{zhang2021varifocalnet}, we design a better mining factor $\gamma$ to obtain more effective gradients from hard cases during training. In detail, we propose a novel adaptive adjusting strategy for $\alpha$ and $\gamma$, concerning the metric on the current convergence situation. Therefore, we define the loss function in the following Eq. \ref{form 3}: all the original settings have been kept, except for our proposed adjusting schedule on $\alpha_{ad}$ and $\gamma_{ad}$. Also, we can easily apply our approach to other extensions of focal loss (i.e., Eq. \ref{form:pqfl} and Eq. \ref{form:vqfl}).

    \begin{align}
        \label{form 3}
        {\rm HPF}(p_i, y_i)=\left\{
        \begin{aligned}
            & -\alpha_{ad} (1-p_i)^{\gamma_{ad}} {\rm log}(p_i), &     &  y_i=1 \\
            & -(1-\alpha_{ad})p_i^{\gamma_{ad}} {\rm log}(1-p_i), &     &  y_i=0.
        \end{aligned}
        \right.
    \end{align}%

    \textbf{Progressive Focus on hard-case mining.} In detail, we design $\gamma_{ad}$ based on the positive samples' prediction quality. Eq. \ref{form 4} defines $\gamma_{ad}$ according to the cross-entropy (CE) loss, naturally corresponding to the KL distance from predictions to targets during training. As shown as follows, $y_i$ is the assigned label through a specific assignment strategy, and $p_i$ indicates the corresponding probability result, after activated by the sigmoid function. Therefore, ${y}_i \cdot {p}_i$ indicates the probability results for all positive samples, and is set as 0 for negative samples. We define $n$ as the number of total samples, and $n_{pos}$ as the number of total positive samples. Besides, the adjusted hyperparameter is clamped within a valid interval $[\gamma - \delta, \gamma + \delta]$ with $\delta$ set as a constant.

    \begin{align}
        \label{form 4}
            \gamma_{ad} = - {\rm log}(\frac{1}{n_{pos}}\sum_{i=1}^{n} {y}_i \cdot {p}_i).
    \end{align}%

    \textbf{Progressive Focus on class-balance.} We set $\alpha_{ad}$ to follow a negative correlation with $\gamma_{ad}$ timely. From our practical experience, $\alpha_{ad}$ should maintain the total gradient strength when $\gamma_{ad}$ progressively changes, keeping the classification loss at a reasonable scale with other losses (e.g., regression loss, centerness loss, etc.). As shown in Eq. \ref{form 5}, $w$ is a constant to calculate $\alpha_{ad}$ from a negative correlation by $\gamma_{ad}$. Besides, we calculate $\alpha_{ad}$ and $\gamma_{ad}$ instantly and do not generate any extra gradient through these two adaptive parameters during training.

    \begin{align}
        \label{form 5}
            \alpha_{ad} = w / \gamma_{ad}.
    \end{align}%

    Therefore, $\gamma_{ad}$ reflects the whole convergence situation with a large value initially and then progressively decreasing with the hard cases reducing. Corresponding to Fig. \ref{fig_3} (b), such an adjusting schedule can initially make the model emphasize hard cases. Then, when the hard and easy cases are not discriminative enough, our approach can gradually increase the distinguishing strength, maintaining the dominance of hard cases throughout the whole training process.

    \subsection{Hierarchical Sampling}\label{sec:hieSample}
    Hierarchical sampling is a practical level-wise approach to generate a set of $\alpha_{ad}$ and $\gamma_{ad}$, without any level-specific settings.
    
    \textbf{Approach details.} As shown in Fig. \ref{fig_1}, we calculate $\alpha_{ad}$ and $\gamma_{ad}$ by Eq. \ref{form 4} and Eq. \ref{form 5}, through only sampling the positive samples' predictions $p_i$ from one level. Then, we calculate the level-wise classification losses by Eq. \ref{form 3} for each level. Finally, we calculate the total classification loss by the mean value of all level-wise classification losses, following Eq. \ref{form 1}.

    \begin{align} 
        \label{form 1}
            {\rm Loss}_{cls} = \frac{1}{L} \sum_{l=1}^L {\rm HPF}_l(P_l, Y_l).
    \end{align}%

    In Eq. \ref{form 1}, $L$ is the number of pyramid levels in the one-stage detector, $P_l$ is the level-wise prediction results generated only from the $l^{th}$ level, and $Y_l$ indicates the assigned labels through a specific target assignment strategy (various in different detectors). Then, ${\rm HPF}_l$ corresponds to the loss of the $l^{th}$ level's predictions, and ${\rm Loss}_{cls}$ is the mean value of the set of ${\rm HPF}_l$.
    In general, hierarchical sampling encourages each pyramid level flexibility to adapt for their relevant target distribution without any level-wise particular setting.
    
    The total optimization procedure with HPF is described in Algorithm \ref{alg:algorithm}. We introduce the whole paradigm as a multi-level optimization form, for easily applied into any multi-level one-stage detector.

	\begin{algorithm}[t]
	\caption{Training with HPF for multi-level one-stage  detectors}
	\label{alg:algorithm}
  	\SetAlgoLined
  	\KwIn{a set of predictions P, a set of corresponding ground truths Y}
  	\KwOut{the total classification loss ${\rm Loss}_{cls}$}
  	split P to subsets ${\rm S}_p={\rm [P_1,P_2,\dots]}$ by each level.
  	
  	split Y to subsets ${\rm S}_y={\rm [Y_1,Y_2,\dots]}$ by each level.
  	
  	\For{${\rm P}_l \in {\rm S}_p$, ${\rm Y}_l \in {\rm S}_y$}{
		calculate $\gamma_{ad}$ by ${\rm y}_i \in {\rm Y}_l$ and ${\rm p}_i \in {\rm P}_l$, according to Eq. \ref{form 4}.
		
		calculate $\alpha_{ad}$ by $\gamma_{ad}$, according to Eq. \ref{form 5}.
		
		calculate ${\rm HPF}_l$ for the $l^{th}$ level by $\gamma_{ad}$ and $\alpha_{ad}$, according to Eq. \ref{form 3}.
	}
	calculate ${\rm Loss}_{cls}$, according to the Eq. \ref{form 1}.
	
	\textbf{return} ${\rm Loss}_{cls}$
	\end{algorithm}

\section{Experiments}\label{sec:exp}
	We perform the experiments on the bounding box detection track of the large-scale benchmark MS COCO \cite{lin2014microsoft}. After providing our settings based on ATSS, we perform a detailed ablation study, including comparisons with other comparable losses, performance evaluation under level discrepancy, components' contributions, and discussion on other settings. Then, we verified our general performance improvement based on \textbf{other 8} representative detectors. Finally, we present the comparisons with the SOTA through a series of scaleable settings. 

    \textbf{Dataset.} Based on MS COCO \cite{lin2014microsoft}, we follow the standard practice in previous works \cite{ren2015faster,tian2019fcos} that set \emph{trainval35k} split (115K images) for training and \emph{minival} split (5K images) as validation. 
        In the ablation study, we list the performance on \emph{minival} split under many conditions as detailed as possible. 
        Then, we upload the prediction results on \emph{test-dev} split (20K images) and evaluate our performance online for comparisons with SOTA. Finally, we follow the experimental setting from our ablation study to verify the general performance improvement.

    \textbf{Network setting.} We keep the architecture and the related model-design settings as default from the original, if not otherwise specified. For example, we initialize our backbone networks with the ImageNet (ILSVC) pre-trained weights. In addition, we also follow the original anchor-related settings (i.e., the definition of anchors, target assignment strategy), maintaining the same target distribution for a fair comparison. For single-scale training, we resize the shorter side to 800 and the longer side to less or equal to 1333, keeping the aspect ratio simultaneously. For multi-scale training, we randomly select the shorter side from 640 to 800, resizing the image by the same aspect ratio as the original.

    \textbf{Optimization.} We set HPF to optimize each level predictions independently but without individual treatment for partial levels. For all levels, we adopt $w$ as $\alpha\cdot\gamma$ (0.5) and $\delta$ as 0.5 for our main results, and provide the related analysis on hyperparameters in the following $\S$\ref{sec:hyper}. During training, we set the initial learning rate to 0.01 and decay it by a factor of 10 at 90k and 120k iterations for the 135k iterations. Meanwhile, we set 0.5 as the weight of centerness loss. Besides, we keep all other original settings, including training the model with stochastic gradient descent (SGD), adopting the same linear warmup schedule in the first 500 iterations, setting weight decay as 1e-4 and momentum as 0.9.

    \subsection{Ablation Study}

	\begin{table}[t]
        \begin{minipage}{0.48\linewidth}
            \footnotesize
            \centering
            \begin{tabular}{l|c|c|c}
                Method  & AP & $\rm AP_{50}$ & $\rm AP_{75}$ \\
                \thickhline
                ATSS w/FL  \cite{zhang2020bridging} & 39.3 & 57.5 & 42.8 \\
                ATSS w/QFL \cite{li2020generalized} & 39.9 & 58.5 & 43.0 \\
                ATSS w/VFL \cite{zhang2021varifocalnet} & 40.1 & 58.5 & 43.4 \\  
                \rowcolor{Gray}
                ATSS w/HPF & \textbf{40.5} & \textbf{59.2} & \textbf{43.8} \\  

            \end{tabular}
            \caption{Comparison of four different losses on COCO \emph{minival}. Besides, HPF can further promote QFL and VFL, shown in $\S$\ref{exp:3}.}
            \label{tab:5}
        \end{minipage}
		\begin{minipage}[h]{0.48\linewidth}
			\centering
			\includegraphics[width=0.9\textwidth]{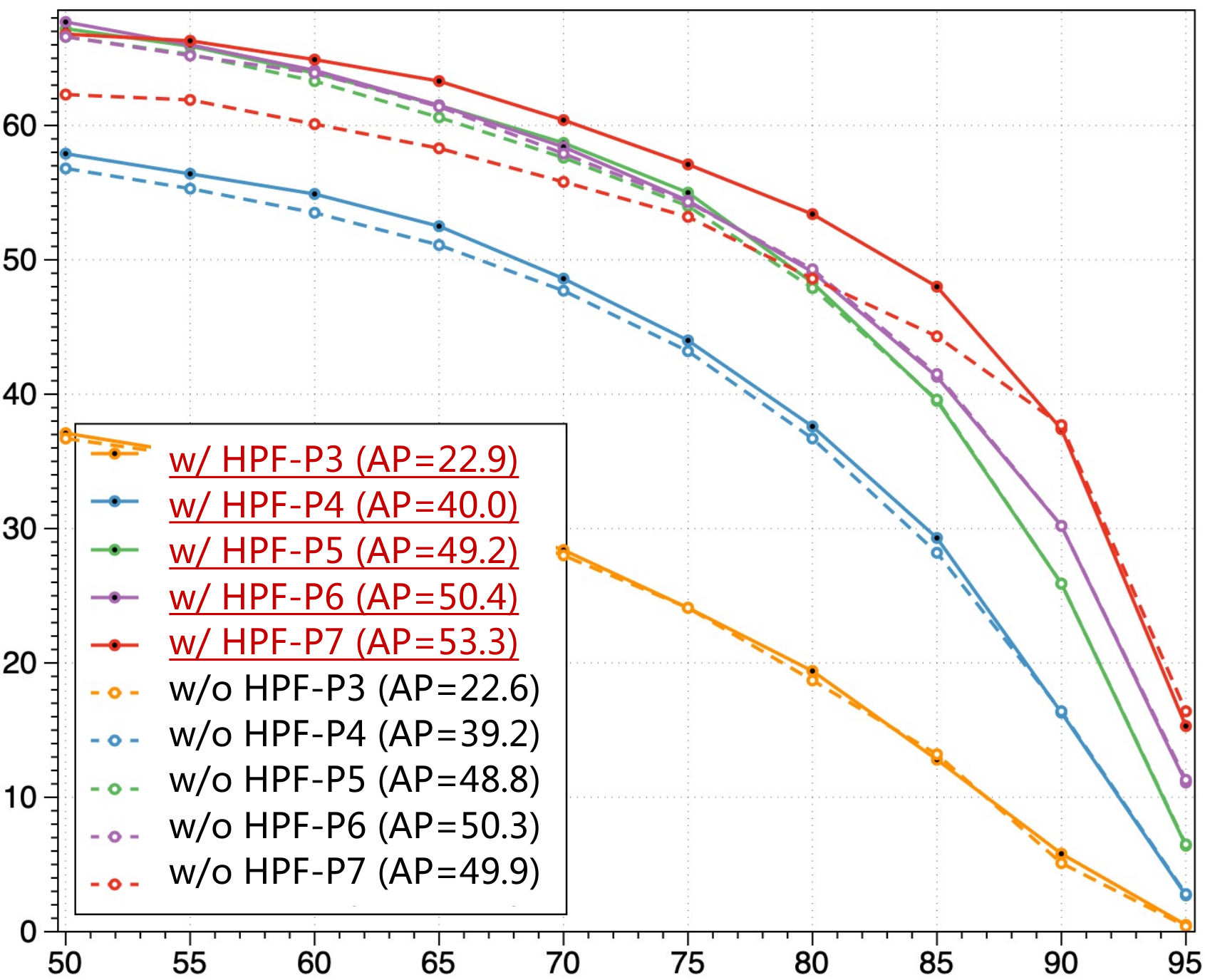}
	        \caption{Visualization of level-wise performance evaluation. We show more quantitative analyses in Tab. \ref{tab:3}}
	        \label{figlwp}
		\end{minipage}%
	\end{table}
	
    \begin{table}[t]
        \footnotesize
        \centering
        \begin{tabular}{l|c|c|c|c|c|c}
            ATSS & w/HPF & P3    & P4    & P5    & P6    & P7    \\
            \thickhline
            \multirow{2}{*}{AP} &   & 22.6 & 39.2 & 48.8 & 50.3 & 49.9 \\
                                & \cellcolor{Gray}{\checkmark} & \cellcolor{Gray}{22.9 (+0.3)} & \cellcolor{Gray}{40.0 (+0.8)} & \cellcolor{Gray}{49.2 (+0.4)} & \cellcolor{Gray}{50.4 (+0.1)} & \cellcolor{Gray}{53.3 (+3.4)} \\
            \hline
            \multirow{2}{*}{$\rm AP_{50}$} &   & 36.7 & 56.8 & 66.6 & 66.6 & 62.3 \\
                                & \cellcolor{Gray}{\checkmark} & \cellcolor{Gray}{37.1 (+0.4)} & \cellcolor{Gray}{57.9 (+1.1)} & \cellcolor{Gray}{67.2 (+0.6)} & \cellcolor{Gray}{67.7 (+1.1)} & \cellcolor{Gray}{66.8 (+4.5)} \\
            \hline
            \multirow{2}{*}{$\rm AP_{75}$} &   & 24.1 & 43.2 & 54.0 & 54.3 & 53.2 \\
                                & \cellcolor{Gray}{\checkmark} & \cellcolor{Gray}{24.1} & \cellcolor{Gray}{44.0 (+0.8)} & \cellcolor{Gray}{55.0 (+1.0)} & \cellcolor{Gray}{54.4 (+0.1)} & \cellcolor{Gray}{57.1 (+3.9)} \\
        \end{tabular}
        \caption{Level-wise analysis. HPF makes performance improvement without trade off, sometimes by a large margin.}
        \label{tab:3}
    \end{table}

    \textbf{Comparison with other advanced losses.} This section first investigates the performance based on ATSS with ResNet-50, where we apply HPF to the original focal loss (according to Algorithm \ref{alg:algorithm}). Table \ref{tab:5} shows our performance improvement compared with recent works.
    
    We evaluate our contribution with two advanced practices. As shown in Tab. \ref{tab:5}, baseline w/FL indicates the original ATSS trained by focal loss, and baseline w/QFL indicates the performance when only applying QFL (quality focal loss) for replacing FL. Similarly, VFL corresponds to the varifocal loss. The performance of all three existing losses is reported from their original works. Therefore, we only report our results based on the \emph{vanilla} focal loss without adopting any improved designs from QFL and GFL.
    
    Based on the same baseline, HPF can directly achieve 40.5 AP on COCO \emph{minival}. Without any network refinement, our approach increases the AP value by 1.2, surpassing the state-of-the-art QFL (39.9 AP reported from \cite{li2020generalized}) and VFL (40.1 AP reported from \cite{zhang2021varifocalnet}). Furthermore, our approach does not conflict with QFL and VFL, as shown in the following $\S$\ref{exp:3}.

    \begin{table}[t]
        \begin{minipage}{0.48\linewidth}
            \footnotesize
            \centering
            \begin{tabular}{l|c|c|c|c|c}
                 & H. S. & P. F. & AP & $\rm AP_{50}$ & $\rm AP_{75}$ \\
                \thickhline
                ATSS & & & 39.3 & 57.5 & 42.8 \\
                ATSS & \checkmark & & $\rm 39.3^{\dag}$ & $\rm 57.5^{\dag}$ & $\rm 42.8^{\dag}$ \\
                ATSS & & \checkmark & 40.1 & 58.5 & 43.6 \\
                \rowcolor{Gray}
                ATSS & \checkmark & \checkmark & \textbf{40.5} & \textbf{59.2} & \textbf{43.8} \\
            \end{tabular}
            \caption{Components analysis. H.S.: hierarchical sampling. P.F.: progressive focus. $\dag$: Directly applying H.S. will not change anything for optimization.}
            \label{tab:abla}
        \end{minipage}
        \begin{minipage}{0.48\linewidth}
            \footnotesize
            \centering
            \begin{tabular}{l|c|c|c|c}
                 & Sampling Approach & AP & $\rm AP_{50}$ & $\rm AP_{75}$    \\
                \thickhline
                HPF & all-level  & 40.1 & 58.5 & 43.6 \\
                HPF & per-sample  & 40.2 & 58.9 & 43.4 \\
				\rowcolor{Gray}
                HPF & level-wise & \textbf{40.5} & \textbf{59.2} & \textbf{43.8} \\
            \end{tabular}
            \caption{Performance evaluation with various sampling approaches.}
            \label{tab:8}
        \end{minipage}
        
    \end{table}

    \textbf{Evaluation under level discrepancy.} Corresponding to $\S$\ref{sec:levelImbalance}, we perform a further level-wise performance comparison with and without HPF. For comparing fairly, we apply our training settings for training the original ATSS (achieve a slight performance improvement), then evaluate the performance level by level according to their relevant subsets. Finally, we summarize the results through AP-IoU curves and a detailed quantitative table: Fig. \ref{figlwp} illustrates the level-wise AP metric under different criteria (with different IoU thresholds for judging hit or not), and Tab. \ref{tab:3} demonstrates detailed quantitative results.
    
    Both shown in Tab. \ref{tab:3} and Fig. \ref{figlwp}, our method can stably outperform the original performance level by level, sometimes surpassing it by large margins (i.e., increasing the AP value of P7 by 3.4). Compared with the original, our approach makes a performance improvement without any level-wise trade-off, providing a better synergy between FPN and multi-level optimization.
    
    \textbf{Individual component contributions.} HPF is a plug-and-play scheme consisting of two key designs, so we also perform further analysis on each component contribution, summarized as Tab. \ref{tab:abla}. We calculate the \emph{unique} $\alpha_{ad}$ and $\gamma_{ad}$ for all five levels when only applying progressive focus, according to the whole positive predictions. When applying progressive focus and hierarchical sampling, we calculate $\alpha_{ad}$ and $\gamma_{ad}$ \emph{level by level}, and optimize each level with their own losses independently. Besides, directly applying hierarchical sampling on focal loss will not change anything for optimization, so we report the performance same to the original focal loss in Tab. \ref{tab:abla}.
    
    According to Tab. \ref{tab:abla}, the original work reported AP values of 39.3. However, with the progressive focus applied only, the AP value increases by 0.8. Then, with hierarchical sampling applied, the AP value further increases by 0.4. These results clearly show that all two designs of HPF both positively impact performance.
    
    \textbf{Exploration with various sampling approaches.} For a further discussion on hierarchical sampling, we delve into the performance impact of different sampling approaches, including sampling from the entire set (all-level), sampling from the level-wise set (level-wise), and sampling for each sample (per-sample), shown as Tab. \ref{tab:8}. In detail, sampling per-sample indicates we calculate pairs of $\alpha_{ad}$ and $\gamma_{ad}$ for each sample during training. Sampling from all-level indicates we utilize the whole positive prediction set for a unique pair of $\alpha_{ad}$ and $\gamma_{ad}$, identical to only applying progressive focus for optimization in Tab. \ref{tab:abla}. Besides, level-wise sampling corresponds to our proposed methods. According to the performance shown in Tab. \ref{tab:8}, hierarchical sampling surpasses the others with a stable margin.
  
    \begin{table}[t]
        \begin{minipage}{0.48\linewidth}
            \footnotesize
            \centering
            \begin{tabular}{c|>{\columncolor{Gray}}ccc|ccc}
                $\delta$ & AP & $\rm AP_{50}$ & $\rm AP_{75}$ & $\rm AP_{s}$ & $\rm AP_{m}$ & $\rm AP_{l}$ \\
                \thickhline
                0.1 & 39.8 & 58.1 & 43.2 & 23.6 & 43.5 & 51.1 \\
                0.25 & 40.3 & 58.6 & 43.7 & 23.9 & 43.7 & 51.9 \\
                \textbf{0.5} & \textbf{40.5} & \textbf{59.2} & \textbf{43.8} & \textbf{24.0} & \textbf{44.1} & \textbf{51.9} \\
                1.0 & 40.4 & 58.8 & 44.0 & 23.4 & 44.6 & 52.0 \\
                2.0 & 39.4 & 57.7 & 42.8 & 22.7 & 44.4 & 50.5 \\
            \end{tabular}
        \caption{Hyperparameter $\delta$ analysis on COCO \emph{minival}}
        \label{tab:delta}
        \end{minipage}
        \begin{minipage}{0.48\linewidth}
        \footnotesize
        \centering
        \begin{tabular}{c|ccc|ccc}
            $w$ & AP & $\rm AP_{50}$ & $\rm AP_{75}$ & $\rm AP_{s}$ & $\rm AP_{m}$ & $\rm AP_{l}$ \\
            \thickhline
            $\rm 0.5^{3}$ & 38.4 & 55.6 & 41.8 & 22.1 & 40.9 & 49.3 \\
            $\rm 0.5^{2}$ & 39.7 & 57.5 & 43.0 & 23.0 & 43.0 & 51.1 \\
            \rowcolor{Gray}
            \textbf{0.5} & \textbf{40.5} & \textbf{59.2} & \textbf{43.8} & \textbf{24.0} & \textbf{44.1} & \textbf{51.9} \\
            $1.0$ & 39.6 & 58.0 & 42.8 & 23.3 & 43.3 & 50.7 \\
            $1.5$ & 0.0 & 0.0 & 0.0 & 0.0 & 0.0 & 0.0 \\
        \end{tabular}

        \caption{Hyperparameter $w$ analysis on COCO \emph{minival}}
        \label{tab:w}
        \end{minipage}
    \end{table}

    \textbf{Hyperparameter Influence.}\label{sec:hyper} We analyze the effect of hyperparameter in Tab. \ref{tab:delta} and Tab. \ref{tab:w}. $w$ is a constant to make $\alpha_{ad}$ follow the trend of $\gamma_{ad}$, and $\delta$ forms a valid interval for the adjusted hyperparameter $\gamma_{ad}$. Based on ATSS with ResNet-50, we show the performance on COCO \emph{minival} in Tab. \ref{tab:delta} when varying $\delta$ from 0.1 to 2.0 with $w$ fixed as 0.5, and Tab. \ref{tab:w} illustrates the performance when varying $w$ from 0.125 to 1.5 with $\delta$ fixed as 0.5 correspondingly. Results in Tab. \ref{tab:delta} show our method is quite robust with different $\delta$. Results in Tab. \ref{tab:w} illustrate that $ \alpha \cdot \gamma$ is the best choice for setting $w$. According to the performance evaluation, $\delta = 0.5$ and $w=0.5$ are adopted for all the rest of the experiments.

    \subsection{Generality of HPF}\label{exp:3}

    To verify our approach's generality, we apply it to some other one-stage detectors, including RetinaNet \cite{lin2017focal}, GFL \cite{li2020generalized}, GFL-v2 \cite{li2021generalized}, RepPoints \cite{yang2019reppoints}, PAA \cite{kim2020probabilistic}, VFNet \cite{zhang2021varifocalnet}, OTA \cite{ge2021ota}, and TOOD \cite{feng2021tood}, then evaluate the performance on \emph{minival} split. All experiments have been performed based on the ResNet-50 backbone. We conducted our experiments by only applying HPF for replacing the original for the detectors trained by the original focal loss, and applied our two key designs on their loss form for the detectors trained by the focal loss extensions.
    
    \textbf{Extension on QFL.} As a part of GFL \cite{li2020generalized}, QFL extends focal loss compatible with continuous space based on the class-aware location quality predictions. Similar to Eq. \ref{form 3}, we apply $\gamma_{ad}$ to replace the original constant $\beta$, and add a balanced factor $\alpha_{ad}$ for the balance between positive and negative samples, shown as following Eq. \ref{form:pqfl}. 

    \begin{align}
        \label{form:pqfl}
        {\rm PF-QFL}(p_i, q_i)=-\lvert q_i-p_i \rvert^{\gamma_{ad}}((1-\alpha_{ad})(1-q_i){\rm log}(1-p_i)+\alpha_{ad}q_i{\rm log}(p_i)).
    \end{align}%
    
    \textbf{Extension on VFL.} For VFL \cite{zhang2021varifocalnet}, we apply $\alpha_{ad}$ and $\gamma_{ad}$ to replace the constant $\alpha$ and $\gamma$, keeping other settings as original.
    
    \begin{align}
        \label{form:vqfl}
        {\rm PF-VFL}(p_i, q_i)=\left\{
        \begin{aligned}
            & -q_i (q_i{\rm log}(p_i) + (1-q_i) {\rm log}(1-p_i)), &     &  q_i>0 \\
            & -\alpha_{ad} p_i^{\gamma_{ad}} {\rm log}(1-p_i), &     &  q_i=0.
        \end{aligned}
        \right.
    \end{align}%

    \begin{table}[t]
        \centering
        \footnotesize
            \begin{tabular}{c|ccc|c|ccc}
                  Method & AP & $\rm AP_{50}$ & $\rm AP_{75}$ & Method & AP & $\rm AP_{50}$ & $\rm AP_{75}$ \\
                \thickhline
                 RetinaNet & 35.7 & 55.0 & 38.5 &  RepPoints & 38.3 & 59.2 & 41.1 \\
                 \rowcolor{Gray}
                 RetinaNet w/HPF & \textbf{36.9} & 56.3 & 39.4 & RepPoints w/HPF & \textbf{38.8} & 59.7 & 41.7\\
                \hline
                 VFNet & 41.6 & 59.5 & 45.0 & PAA & 41.1 & 59.4 & 44.3\\
                 \rowcolor{Gray}
                 VFNet w/HPF & \textbf{41.8} & 60.0 & 45.2 & PAA w/HPF & \textbf{41.3} & 59.2 & 45.2 \\
                \hline
                 GFL & 40.2 & 58.4 & 43.3 & OTA & 40.7 & 58.4 & 44.3\\
                 \rowcolor{Gray}
                 GFL w/HPF & \textbf{40.7} & 58.5 & 44.0 & OTA w/HPF & \textbf{40.9} & 59.6 & 43.8\\
                \hline
                 GFLv2 & 41.0 & 58.5 & 45.0 & TOOD & 42.4 & 59.5 & 46.1 \\
                 \rowcolor{Gray}
                 GFLv2 w/HPF & \textbf{41.2} & 58.8 & 44.7 & TOOD w/HPF & \textbf{42.7} & 60.2 & 46.1 \\            \end{tabular}
        \caption{Generality on other 8 one-stage detectors, including various assignment strategies, various anchor definitions, and even other extensions of focal loss.}
        \label{tab:9}
    \end{table}
    
    \textbf{Results analysis.} According to Tab. \ref{tab:9}, we can observe a general performance improvement. For RetinaNet, PAA, OTA, and RepPoints, the original works reported AP values of 35.7, 41.1, 40.7, 38.3 for the ResNet-50 backbone. However, with our approach only, the AP value increases by 1.2, 0.2, 0.2, 0.5, respectively, indicating our stable improvement on both anchor-based and anchor-free detectors. For GFL, GFL-v2, VFNet, and TOOD, the AP value is increased to 40.7, 41.2, 41.8, and 42.7, respectively, indicating that our approach can work well with other extensions on focal loss. These results clearly show that our method can cooperate well with recent advances.

    \subsection{Comparison with State of the Art}
    \begin{table*}[t]
        \scriptsize
        \centering
        \begin{tabular}{l|l|c|c|c|ccc|ccc}
        Method & Backbone & Size & Epoch & $\rm MS_{train}$ & AP & $\rm AP_{50}$ & $\rm AP_{75}$ & $\rm AP_{S}$ & $\rm AP_{M}$ & $\rm AP_{L}$ \\
        
        \thickhline
        \emph{Comparison with ATSS:} & & & & & & & & & \\
        ATSS \cite{zhang2020bridging} & R-101-DCN & \underline{800} & 24 & \checkmark & 46.3 & 64.7 & 50.4 & 27.7 & 49.8 & 58.4 \\
        \rowcolor{Gray}
        ATSS w/HPF & R-101-DCN & \underline{800} & 18 & \checkmark & \textbf{47.7} & \textbf{66.9} & \textbf{52.1} & \textbf{29.1} & \textbf{51.0} & \textbf{59.7} \\
        ATSS & X-101-DCN & \underline{800} & 24 & \checkmark & 47.7 & 66.5 & 51.9 & 29.7 & 50.8 & 59.4 \\
        \rowcolor{Gray}
        ATSS w/HPF & X-101-DCN & \underline{800} & 18 & \checkmark & \textbf{49.1} & \textbf{68.5} & \textbf{53.6} & \textbf{30.8} & \textbf{52.6} & \textbf{61.1} \\
        \hline
        \hline
        \emph{Comparison with SOTA:} & & & & & & & & & \\
        Faster R-CNN w/FPN \cite{lin2017feature} & R-101 & \underline{800} & 24 & & 36.2 & 59.1 & 39.0 & 18.2 & 39.0 & 48.2 \\
        Cascade R-CNN \cite{cai2018cascade} & R-101 & \underline{800} & 18 & & 42.8 & 62.1 & 46.3 & 23.7 & 45.5 & 55.2 \\
        CenterNet2* \cite{zhou2021probabilistic} & X-101-DCN & \underline{800} & 24 & \checkmark & 50.2 & 68.0 & 55.0 & 31.2 & 53.5 & 63.6 \\
        
        \hline
        CornerNet \cite{law2018cornernet} & Hg-104 & 512 & 200 & \checkmark & 42.2 & 57.8 & 45.2 & 20.7 & 44.8 & 56.6 \\
        FASF* \cite{zhu2019feature} & X-101 & \underline{800} & 18 & \checkmark & 44.6 & 65.2 & 48.6 & 29.7 & 47.1 & 54.6 \\
        CenterNet* \cite{zhou2019objects} & Hg-104 & 512 & 190 & \checkmark & 45.1 & 63.9 & 49.3 & 26.6 & 47.1 & 57.7 \\
        FCOS* \cite{tian2019fcos} & X-101 & \underline{800} & 24 & \checkmark & 50.4 & 68.9 & 55.0 & 33.2 & 53.0 & 62.7 \\
        ATSS* \cite{zhang2020bridging} & X-101-DCN & \underline{800} & 24 & \checkmark & 50.7 & 68.9 & 56.3 & 33.2 & 52.9 & 62.4 \\
        OTA* \cite{ge2021ota} & X-101-DCN & \underline{800} & 24 & \checkmark & 51.5 & 68.6 & 57.1 & 34.1 & 53.7 & 64.1 \\
        VFNet \cite{zhang2021varifocalnet} & R2-101-DCN & \underline{960} & 24 & \checkmark & 51.3 & 69.7 & 55.8 & 31.9 & 54.7 & 64.4 \\
        TOOD* \cite{feng2021tood} & X-101-DCN & \underline{800} & 24 & \checkmark & 51.1 & 69.4 & 55.5 & 31.9 & 54.1 & 63.7 \\
        Reppoints-v2* \cite{chen2020reppoints} & X-101 & \underline{800} & 24 & \checkmark & 52.1 & 70.1 & 57.5 & 34.5 & 54.6 & 63.6 \\
        GFL-v2* \cite{li2021generalized} & R2-101-DCN & \underline{800} & 24 & \checkmark & 53.3 & 70.9 & 59.2 & 35.7 & 56.1 & 65.6 \\
        PAA* \cite{kim2020probabilistic} & X-101-DCN & \underline{800} & 24 & \checkmark & 53.5 & 71.6 & 59.1 & 36.0 & 56.3 & 66.9 \\
        EfficientDet-D7x \cite{tan2020efficientdet} & EffNet-B7 & 1536 & 600 & \checkmark & 55.1 & 74.3 & 59.9 & 37.2 & 57.9 & \textbf{68.0} \\
        \rowcolor{Gray}
        ATSS w/HPF* & Swin-L-22K & \underline{960} & 24 & \checkmark & \textbf{55.1} & \textbf{74.2} & \textbf{60.7} & \textbf{38.1} & \textbf{58.6} & 66.8 \\
        \hline
        \hline
        \emph{With other settings:} & & & & & & & & & \\
        ATSS w/HPF & R2-101-DCN & \underline{960} & 24 & \checkmark & 50.3 & 69.5 & 54.9 & 31.8 & 53.8 & 63.1 \\
        
        ATSS w/HPF & Swin-S-1K & \underline{960} & 24 & \checkmark & 50.3 & 70.0 & 54.9 & 32.0 & 53.6 & 63.1 \\
        
        ATSS w/HPF & Swin-B-22K & \underline{960} & 24 & \checkmark & 51.9 & 71.6 & 56.6 & 33.4 & 55.4 & 65.0 \\
        
        ATSS w/HPF & Swin-L-22K & \underline{960} & 24 & \checkmark & 53.1 & 72.7 & 58.0 & 34.9 & 56.5 & 66.4 \\

        \end{tabular}
        \caption{Performance comparison on COCO \emph{test-dev}. As we hope HPF can serve as a \emph{common-used freebie}, we choose the common-used ATSS as our based models. 'R': ResNet, 'X': ResNeXt-64x4d, 'R2': Res2Net, 'Hg': Hourglass, 'EffNet': EfficientNet, 'DCN': DCN-v2 \cite{zhu2019deformable}, \underline{960}: resize the shorter side to 960 and the longer less or equal to 1333, $\rm MS_{train}$: scale range 1333$\times$[640:800] for \underline{800}, and 1333$\times$[480:960] for \underline{960}.}
        \label{tab:4}
    \end{table*}

    As HPF is a plug-and-play freebie for general performance improvement, we select the original ATSS as our base model and provide a series of \emph{common-used} but still powerful results without any bells and whistles. Besides, our training settings are typical and easy to follow as well.
    
    We evaluate models on COCO \emph{test-dev} and compare recent state-of-the-art models, mainly on one-stage detectors. Tab. \ref{tab:4} lists our results and the performance of some popular models over recent years. Here we combined our method with more advanced works, more intensive computation, and the best hyperparameter settings to achieve a more competitive final result. 
    Detailedly, we compare our methods with the original ATSS on different backbones, including ResNet-101, ResNeXt-101, Res2Net-101, and the series of Swin Transformer. The multi-scale training strategy and the deformable convolution layer (DCN-v2) are selectively adopted for a fair comparison. For Res2Net-101 and Swin series, the training epoch is up to 24 to ensure convergence. For training with Swin Transformer, 4 patches and 7 windows are kept default as its inner structure settings. '1K' in $2^{nd}$ column means the backbone is pre-trained from the ImageNet-1K, and '22K' indicates the ImageNet-22K correspondingly. Multi-scale testing is adopted on our best single model. We trained all our experiments on 8 Tesla-V100-16GB GPUs, except that the experiments with Swin-L-22K on 8 Tesla-P40-24GB GPUs with NVIDIA-Apex toolkit (utilizing automatic mixed precision for GPU memory saving).

    Compared with the high-performance detectors through long training epochs (600 epochs), large resolution (1536$\times$1536), and enormous computing resources (128 TPUv3), our model achieves a comparable performance of \textbf{55.1 AP} with a basic laboratory setting (24 epochs, 1333$\times$960 resolution, 8 Tesla-24G GPUs), achieving comparable performance with 5$\%$ GPU usage and 4$\%$ epochs.

\section{Conclusions}
    We confirm that gradients drift and level discrepancy problems commonly exist in training one-stage detectors. Based on these problems, we develop a novel plug-and-play scheme named HPF. Through extensive experiments and analyses on the COCO dataset, we verify that HPF is effective and universality in improving one-stage detector performance. We hope HPF can serve as a common-used freebie in one-stage detectors for better multi-level predictions.



\bibliographystyle{splncs}
\bibliography{eccv2016submission}

\begin{thebibliography}{10}

\bibitem{zou2019object}
Zou, Z., Shi, Z., Guo, Y., Ye, J.:
\newblock Object detection in 20 years: A survey.
\newblock arXiv preprint arXiv:1905.05055 (2019)

\bibitem{liu2016ssd}
Liu, W., Anguelov, D., Erhan, D., Szegedy, C., Reed, S., Fu, C.Y., Berg, A.C.:
\newblock Ssd: Single shot multibox detector.
\newblock In: European conference on computer vision, Springer (2016)  21--37

\bibitem{lin2017focal}
Lin, T.Y., Goyal, P., Girshick, R., He, K., Doll{\'a}r, P.:
\newblock Focal loss for dense object detection.
\newblock In: Proceedings of the IEEE international conference on computer
  vision. (2017)  2980--2988

\bibitem{zhang2021varifocalnet}
Zhang, H., Wang, Y., Dayoub, F., Sunderhauf, N.:
\newblock Varifocalnet: An iou-aware dense object detector.
\newblock In: Proceedings of the IEEE/CVF Conference on Computer Vision and
  Pattern Recognition. (2021)  8514--8523

\bibitem{rezatofighi2019generalized}
Rezatofighi, H., Tsoi, N., Gwak, J., Sadeghian, A., Reid, I., Savarese, S.:
\newblock Generalized intersection over union: A metric and a loss for bounding
  box regression.
\newblock In: Proceedings of the IEEE/CVF Conference on Computer Vision and
  Pattern Recognition. (2019)  658--666

\bibitem{tian2019fcos}
Tian, Z., Shen, C., Chen, H., He, T.:
\newblock Fcos: Fully convolutional one-stage object detection.
\newblock In: Proceedings of the IEEE/CVF international conference on computer
  vision. (2019)  9627--9636

\bibitem{zhang2020bridging}
Zhang, S., Chi, C., Yao, Y., Lei, Z., Li, S.Z.:
\newblock Bridging the gap between anchor-based and anchor-free detection via
  adaptive training sample selection.
\newblock In: Proceedings of the IEEE/CVF conference on computer vision and
  pattern recognition. (2020)  9759--9768

\bibitem{lin2017feature}
Lin, T.Y., Doll{\'a}r, P., Girshick, R., He, K., Hariharan, B., Belongie, S.:
\newblock Feature pyramid networks for object detection.
\newblock In: Proceedings of the IEEE conference on computer vision and pattern
  recognition. (2017)  2117--2125

\bibitem{zhu2019feature}
Zhu, C., He, Y., Savvides, M.:
\newblock Feature selective anchor-free module for single-shot object
  detection.
\newblock In: Proceedings of the IEEE/CVF Conference on Computer Vision and
  Pattern Recognition. (2019)  840--849

\bibitem{tan2020efficientdet}
Tan, M., Pang, R., Le, Q.V.:
\newblock Efficientdet: Scalable and efficient object detection.
\newblock In: Proceedings of the IEEE/CVF conference on computer vision and
  pattern recognition. (2020)  10781--10790

\bibitem{kong2020foveabox}
Kong, T., Sun, F., Liu, H., Jiang, Y., Li, L., Shi, J.:
\newblock Foveabox: Beyound anchor-based object detection.
\newblock IEEE Transactions on Image Processing \textbf{29} (2020)  7389--7398

\bibitem{chen2021you}
Chen, Q., Wang, Y., Yang, T., Zhang, X., Cheng, J., Sun, J.:
\newblock You only look one-level feature.
\newblock In: Proceedings of the IEEE/CVF Conference on Computer Vision and
  Pattern Recognition. (2021)  13039--13048

\bibitem{cai2018cascade}
Cai, Z., Vasconcelos, N.:
\newblock Cascade r-cnn: Delving into high quality object detection.
\newblock In: Proceedings of the IEEE conference on computer vision and pattern
  recognition. (2018)  6154--6162

\bibitem{ren2015faster}
Ren, S., He, K., Girshick, R., Sun, J.:
\newblock Faster r-cnn: Towards real-time object detection with region proposal
  networks.
\newblock Advances in neural information processing systems \textbf{28} (2015)
  91--99

\bibitem{pang2019libra}
Pang, J., Chen, K., Shi, J., Feng, H., Ouyang, W., Lin, D.:
\newblock Libra r-cnn: Towards balanced learning for object detection.
\newblock In: Proceedings of the IEEE/CVF Conference on Computer Vision and
  Pattern Recognition. (2019)  821--830

\bibitem{chen2019hybrid}
Chen, K., Pang, J., Wang, J., Xiong, Y., Li, X., Sun, S., Feng, W., Liu, Z.,
  Shi, J., Ouyang, W.,  et~al.:
\newblock Hybrid task cascade for instance segmentation.
\newblock In: Proceedings of the IEEE/CVF Conference on Computer Vision and
  Pattern Recognition. (2019)  4974--4983

\bibitem{dalal2005histograms}
Dalal, N., Triggs, B.:
\newblock Histograms of oriented gradients for human detection.
\newblock In: 2005 IEEE computer society conference on computer vision and
  pattern recognition (CVPR'05). Volume~1., Ieee (2005)  886--893

\bibitem{felzenszwalb2010cascade}
Felzenszwalb, P.F., Girshick, R.B., McAllester, D.:
\newblock Cascade object detection with deformable part models.
\newblock In: 2010 IEEE Computer society conference on computer vision and
  pattern recognition, Ieee (2010)  2241--2248

\bibitem{girshick2014rich}
Girshick, R., Donahue, J., Darrell, T., Malik, J.:
\newblock Rich feature hierarchies for accurate object detection and semantic
  segmentation.
\newblock In: Proceedings of the IEEE conference on computer vision and pattern
  recognition. (2014)  580--587

\bibitem{girshick2015fast}
Girshick, R.:
\newblock Fast r-cnn.
\newblock In: Proceedings of the IEEE international conference on computer
  vision. (2015)  1440--1448

\bibitem{song2020revisiting}
Song, G., Liu, Y., Wang, X.:
\newblock Revisiting the sibling head in object detector.
\newblock In: Proceedings of the IEEE/CVF Conference on Computer Vision and
  Pattern Recognition. (2020)  11563--11572

\bibitem{zhu2021cpm}
Zhu, B., Song, Q., Yang, L., Wang, Z., Liu, C., Hu, M.:
\newblock Cpm r-cnn: Calibrating point-guided misalignment in object detection.
\newblock In: Proceedings of the IEEE/CVF Winter Conference on Applications of
  Computer Vision. (2021)  3248--3257

\bibitem{jiang2018acquisition}
Jiang, B., Luo, R., Mao, J., Xiao, T., Jiang, Y.:
\newblock Acquisition of localization confidence for accurate object detection.
\newblock In: Proceedings of the European conference on computer vision (ECCV).
  (2018)  784--799

\bibitem{redmon2017yolo9000}
Redmon, J., Farhadi, A.:
\newblock Yolo9000: better, faster, stronger.
\newblock In: Proceedings of the IEEE conference on computer vision and pattern
  recognition. (2017)  7263--7271

\bibitem{zhou2019objects}
Zhou, X., Wang, D., Kr{\"a}henb{\"u}hl, P.:
\newblock Objects as points.
\newblock arXiv preprint arXiv:1904.07850 (2019)

\bibitem{li2019gradient}
Li, B., Liu, Y., Wang, X.:
\newblock Gradient harmonized single-stage detector.
\newblock In: Proceedings of the AAAI Conference on Artificial Intelligence.
  (2019)

\bibitem{li2020generalized}
Li, X., Wang, W., Wu, L., Chen, S., Hu, X., Li, J., Tang, J., Yang, J.:
\newblock Generalized focal loss: Learning qualified and distributed bounding
  boxes for dense object detection.
\newblock arXiv preprint arXiv:2006.04388 (2020)

\bibitem{li2021generalized}
Li, X., Wang, W., Hu, X., Li, J., Tang, J., Yang, J.:
\newblock Generalized focal loss v2: Learning reliable localization quality
  estimation for dense object detection.
\newblock In: Proceedings of the IEEE/CVF Conference on Computer Vision and
  Pattern Recognition. (2021)  11632--11641

\bibitem{ke2020multiple}
Ke, W., Zhang, T., Huang, Z., Ye, Q., Liu, J., Huang, D.:
\newblock Multiple anchor learning for visual object detection.
\newblock In: Proceedings of the IEEE/CVF Conference on Computer Vision and
  Pattern Recognition. (2020)  10206--10215

\bibitem{zhu2020autoassign}
Zhu, B., Wang, J., Jiang, Z., Zong, F., Liu, S., Li, Z., Sun, J.:
\newblock Autoassign: Differentiable label assignment for dense object
  detection.
\newblock arXiv preprint arXiv:2007.03496 (2020)

\bibitem{kim2020probabilistic}
Kim, K., Lee, H.S.:
\newblock Probabilistic anchor assignment with iou prediction for object
  detection.
\newblock In: Computer Vision--ECCV 2020: 16th European Conference, Glasgow,
  UK, August 23--28, 2020, Proceedings, Part XXV 16, Springer (2020)  355--371

\bibitem{ge2021ota}
Ge, Z., Liu, S., Li, Z., Yoshie, O., Sun, J.:
\newblock Ota: Optimal transport assignment for object detection.
\newblock In: Proceedings of the IEEE/CVF Conference on Computer Vision and
  Pattern Recognition. (2021)  303--312

\bibitem{shrivastava2016training}
Shrivastava, A., Gupta, A., Girshick, R.:
\newblock Training region-based object detectors with online hard example
  mining.
\newblock In: Proceedings of the IEEE conference on computer vision and pattern
  recognition. (2016)  761--769

\bibitem{lin2014microsoft}
Lin, T.Y., Maire, M., Belongie, S., Hays, J., Perona, P., Ramanan, D.,
  Doll{\'a}r, P., Zitnick, C.L.:
\newblock Microsoft coco: Common objects in context.
\newblock In: European conference on computer vision, Springer (2014)  740--755

\bibitem{yang2019reppoints}
Yang, Z., Liu, S., Hu, H., Wang, L., Lin, S.:
\newblock Reppoints: Point set representation for object detection.
\newblock In: Proceedings of the IEEE/CVF International Conference on Computer
  Vision. (2019)  9657--9666

\bibitem{feng2021tood}
Feng, C., Zhong, Y., Gao, Y., Scott, M.R., Huang, W.:
\newblock Tood: Task-aligned one-stage object detection.
\newblock In: 2021 IEEE/CVF International Conference on Computer Vision (ICCV),
  IEEE Computer Society (2021)  3490--3499

\bibitem{zhou2021probabilistic}
Zhou, X., Koltun, V., Kr{\"a}henb{\"u}hl, P.:
\newblock Probabilistic two-stage detection.
\newblock arXiv preprint arXiv:2103.07461 (2021)

\bibitem{law2018cornernet}
Law, H., Deng, J.:
\newblock Cornernet: Detecting objects as paired keypoints.
\newblock In: Proceedings of the European conference on computer vision (ECCV).
  (2018)  734--750

\bibitem{chen2020reppoints}
Chen, Y., Zhang, Z., Cao, Y., Wang, L., Lin, S., Hu, H.:
\newblock Reppoints v2: Verification meets regression for object detection.
\newblock Advances in Neural Information Processing Systems \textbf{33} (2020)
  5621--5631

\bibitem{zhu2019deformable}
Zhu, X., Hu, H., Lin, S., Dai, J.:
\newblock Deformable convnets v2: More deformable, better results.
\newblock In: Proceedings of the IEEE/CVF Conference on Computer Vision and
  Pattern Recognition. (2019)  9308--9316

\end{thebibliography}
\end{document}